 \documentclass[final,3p,times]{elsarticle}

\usepackage{amssymb}
\usepackage{lineno,hyperref}

\journal{Physics of Life Reviews}

\begin{document}

\begin{frontmatter}



\title{A perspective on the advancement of natural language processing tasks via topological analysis of complex networks \\
Comment on ``Approaching human language with complex networks'' by Cong and Liu}


\author{Diego Raphael Amancio\footnote{E-mail:diego.raphael@gmail.com, diego@icmc.usp.br}}

\address{
Institute of Mathematical and Computer Sciences \\
University of S\~ao Paulo \\
S\~ao Carlos, S\~ao Paulo, Brazil}







\end{frontmatter}


\section*{}
\label{}

 Concepts and methods of complex networks have been applied to probe the properties of a myriad of real systems~\cite{app}. The finding that written texts modeled as graphs share several properties of other completely different real systems has inspired the study of language as a complex system~\cite{evolving}. Actually, language can be  represented as a complex network in its several levels of complexity. As a consequence, morphological, syntactical and semantical properties have been employed in the construction of linguistic networks~\cite{largescale}. Even the character level has been useful to unfold particular patterns~\cite{china,japan}. In the review by Cong and Liu~\cite{liucong}, the authors emphasize the need to use the topological information of complex networks modeling the various spheres of the language to better understand its origins, evolution and organization. In addition, the authors cite the use of networks in applications aiming at holistic typology and stylistic variations. In this context, I will discuss some possible directions that could be followed in future research directed towards the understanding of language via topological characterization of complex linguistic networks. In addition, I will comment the use of network models for language processing applications. Additional prospects for future practical research lines will also be discussed in this comment.

The topological analysis of complex textual networks has been widely studied in the recent years. As for co-occurrence networks of characters, it was possible to verify that they follow the scale-free and small-world features~\cite{china}.
Co-occurrence networks of words (or adjacency networks) have accounted for most of the models tackling textual applications. In special, they have been more prevalent than syntactical networks because they represent a simplified representation of the complex syntactical analysis~\cite{Cancho01thesmall,probing}, as most of the syntactical links occur between neighboring words. Despite its outward simplicity, co-occurrence networks have proven useful in many applications, such as in authorship recognition~\cite{comparing}, extractive summarization~\cite{extractive,lantiq,graphsum}, stylistic identification~\cite{identification} and part-of-speech tagging~\cite{chaos}.
Furthermore, such representation has also been useful in the analysis of the complexity~\cite{sanda} and quality of texts~\cite{strong}. Unfortunately, a major problem arising from the analyses performed with co-occurrence networks is the difficulty to provide a rigorous interpretation of the factors accounting for the success of the model. Therefore, future investigations should pursue a better interpretation at the network level aiming at the understanding of the fundamental properties of the language.
Most importantly, it is clear from some recent studies~\cite{probing,comparing} that novel topological measurements should be introduced to capture  a wider range of linguistic features.

Many of the applications relying on network analysis outperform other traditional shallow strategies in natural language processing (see e.g. the extractive summarization task~\cite{extractive,lantiq}). However, when deep analyzes are performed, network-based strategies usually do not perform better than other techniques making extensive use of semantic resources and tools. In order to improve the performance of network-based applications, I suggest a twofold research line: (i) the introduction of measurements consistent with the nature of the problem; and (ii) the combination of topological strategies with other traditional natural language processing methods. More specifically, in (i), I propose the conception of measurements  that are able to capture semantic aspects, since the topological measurements of co-occurrence networks capture mostly syntactic factors~\cite{probing}. Although such networks have proved useful in some semantical-dependent tasks (see e.g. a topological approach to word sense disambiguation in~\cite{unveiling}), I believe that the creation of novel semantic-based measurements would improve the state of the art. Alternative forms to create the network could also be useful to grasp semantical features hidden in the topological space.
In (ii), I suggest, for example, the introduction of a hybrid classifier that could consider both linguistic (deeper linguistic processing~\cite{deeper}) and topological attributes at the same time in a hybrid way. Examples of combinations of distinct strategies are described in~\cite{comparing},~\cite{highorder} and~\cite{topcol}.

In sum, the network framework has proven applicable to understand the properties of the language and its applications, especially those related to the textual classification in several levels. Despite the limitations imposed by the restrict understanding of the mechanisms behind the classification, it is worth noting that the such representation  remains entirely generic, being therefore useful to many tasks as well as for analyzing the evolution of languages, cultures and emotional trends. For this reason, I believe that the
use of complex networks in both practical and theoretical investigations shall yield novels insights into the mechanisms behind the language.

\section*{Acknowledgements}

I acknowledge the financial support from S\~ao Paulo Research Foundation (FAPESP grant number 13/06717-4).





\end{document}